\documentclass[letterpaper, 10 pt, conference]{ieeeconf}  

\usepackage{amsmath}
\usepackage{amsfonts}
\usepackage{graphicx}
\usepackage{subcaption}
\usepackage{wrapfig}
\usepackage{booktabs}
\usepackage{multirow}
\usepackage{threeparttable}
\usepackage{float}
\usepackage{balance}
\usepackage{pgfplots}
\pgfplotsset{compat=newest}

\IEEEoverridecommandlockouts                              

\title{\LARGE \bf
Physically-Based Lighting Generation for Robotic Manipulation}

\author{
Shutong Jin$^{1*}$, Lezhong Wang$^{2*}$, Ben Temming$^1$, Florian T. Pokorny$^1$ \thanks{$*$Equal contribution. $^1$KTH Royal Institute of Technology, 
        $^2$Technical University of Denmark.
        {\tt\{shutong, fpokorny\}@kth.se}. This work was partially supported by the Wallenberg AI, Autonomous Systems and Software Program (WASP) funded by the Knut and Alice Wallenberg Foundation. The computations were enabled by the supercomputing resource Berzelius provided by the National Supercomputer Centre at Linköping University and the Knut and Alice Wallenberg Foundation, Sweden.%
        }
}

\begin{document}

\maketitle
\thispagestyle{empty}
\pagestyle{empty}

\begin{abstract}
In this paper, we propose the first framework that leverages physically-based inverse rendering for novel lighting generation on existing real-world human demonstrations of robotic manipulation tasks.
Specifically, inverse rendering decomposes the first frame in each demonstration into geometric (surface normal, depth) and material (albedo, roughness, metallic) properties, which are then used to render appearance changes under different lighting sources. 
To improve efficiency and maintain consistency across each generated sequence, we fine-tune Stable Video Diffusion on robot execution videos for temporal lighting propagation. 
We evaluate our framework by measuring the visual quality of the generated sequences, assessing its effectiveness in improving the imitation learning policy performance (38.75\%) under six unseen real-world lighting conditions, and conduct ablation studies on individual modules of the proposed framework.
We further showcase three downstream applications enabled by the proposed framework: background generation, object texture generation and distractor positioning.
The code for the framework will be made publicly available.
\end{abstract}


\section{Introduction}
Imitation learning from large-scale human demonstrations has proven to be an effective approach to deploying robotic manipulation tasks~\cite{black2024pi_0}. 
Yet collecting such data is costly, as it often needs to cover both diverse motor skills and varied visual appearances~\cite{yu2023scaling, xie2024decomposing}. 
In a fixed object-environment setup, achieving reliable policy performance on a single skill typically requires a proficient operator to repeat the task about 200 times using specialized teleoperation devices~\cite{mandlekar2023mimicgen}. 
To ensure broader robustness, dataset construction further incorporates visual variations such as object texture, background, and distractors~\cite{walke2023bridgedata, lin2024data}. 
Capturing these variations requires repeating the data collection process for each factor, making it especially costly to achieve sufficient visual coverage for every skill in real-world settings.

In response, substantial effort has been devoted to synthesizing object texture and background variations~\cite{yu2023scaling, mandi2022cacti, chen2023genaug, teoh2024green, yuan2025roboengine}.
Inpainting techniques based on generative modeling~\cite{chen2023genaug} and physical tools such as greenscreens~\cite{teoh2024green} are employed, yielding promising results while reducing the need for additional data collection.
By comparison, lighting, another pervasive and highly dynamic factor in real-world settings, has received little attention.
Even in the relatively controlled indoor environments, lighting can vary significantly with artificial sources such as lamps or daylight through windows. 
In outdoor environments, the variability is far greater, as field robots may be exposed to natural lighting changes such as those caused by the position of the sun or changes in cloud cover.
Such variability points to the need for training data under different lighting.
However, given the existing cost of real-world data collection, expanding datasets to cover each skill under lighting variations only compounds the challenge.

\begin{figure}[t]
    \centering
  \begin{subfigure}[h]{0.12\textwidth}
    \centering
    \text{Real}\\[0.3em]
    \includegraphics[width=\textwidth]{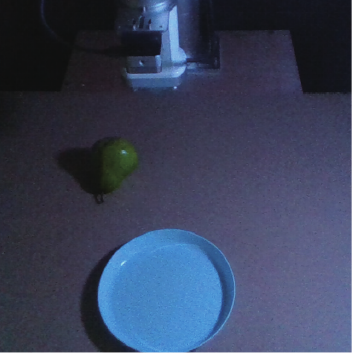}
    \vspace{-1.8 em}
    \caption{}
    \label{fig:firstA}
  \end{subfigure}
  \hspace{1em}
  \begin{subfigure}[h]{0.3\textwidth}
    \centering
    \makebox[0.85\linewidth][r]{Generated}\\[0.3em]
    \includegraphics[width=\textwidth]{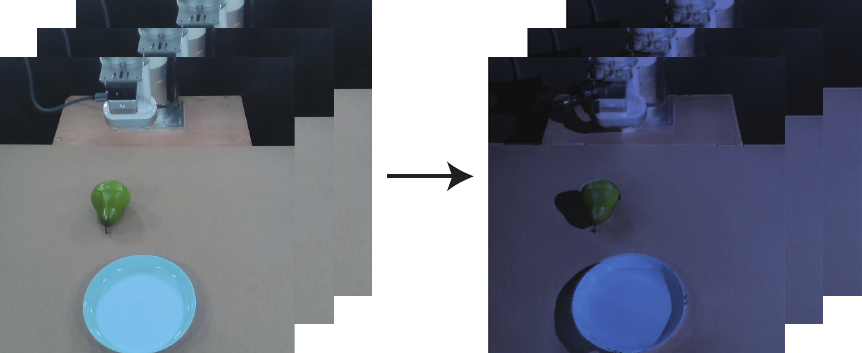}
    \vspace{-1.8 em}
    \caption{}
    \label{fig:firstB}
  \end{subfigure}
    \vspace{-0.5em}
    \caption{\small(a) Example of recursive effect of lighting: when a lamp is turned off, all scene components appear darker (global impact), while nearby components alter one another’s appearance by casting shadows (local impact). (b) Left: existing real-world human demonstration. Right: demonstration relit by our method. Our framework generates novel lighting for real-world demonstrations to approximate scenes under unseen lighting conditions.}
    \label{fig:first}
    \vspace{-1em}
\end{figure}

Meanwhile, lighting poses a unique modeling challenge.
As light travels from its source, it scatters and reflects off all scene components before ultimately being captured by a camera.  Consequently, variations in lighting alter the appearance of the entire scene and influence the training process that depends on camera observations. This recursive property of lighting makes it extremely difficult to synthesize in robotic scenes.
One example is shown in \textit{Fig.}~\ref{fig:firstA}: when a lamp is turned off, all scene components appear darker (global impact), while nearby components alter one another’s appearance by casting shadows (local impact).
Such behaviors may also help explain why many studies report that trained policies are highly vulnerable to lighting variations~\cite{xie2024decomposing, lin2024data, xing2021kitchenshift}.
Taken together, these bring us to the question:

\textit{Can we generate lighting that approximates real-world variations to reduce costly data collection for robotic manipulation?}

To tackle this, we propose \textbf{\textit{RoLight}}, the first framework that leverages physically-based inverse rendering for \textbf{ro}botic scene \textbf{light}ing generation.
Inverse rendering is introduced for explicit modeling of geometric and material information in existing real-world human demonstrations for simulating accurate light-material interactions.
Our contributions are fourfold:
\begin{itemize}
    \item \textbf{Modular Integration. }
    We adopt an inverse rendering module~\cite{wang2025materialist} to decompose a single demonstration frame into geometric and material properties, which are then used by a rendering module to simulate new lighting on the decomposed properties.
    
    \item \textbf{Domain-Adapted Stable Video Diffusion. }
    To improve efficiency and maintain consistent generation across consecutive frames of each demonstration, we fine-tune Stable Video Diffusion (SVD)~\cite{blattmann2023stable} for temporal lighting propagation.  Fine-tuning data includes synthetic robot execution videos from Factor World~\cite{xie2024decomposing} with varied lighting across tasks, and real-world execution videos from RoboNet~\cite{dasari2019robonet} with visual degradation for radiometric accuracy.
    
    \item \textbf{Real-world Evaluation. }  
    We validate our framework by assessing structural and temporal consistency in generated demonstrations, and through real-world experiments on a 7-DoF robot with an embodiment unseen during SVD fine-tuning. Under six varied lighting conditions, our method improves the imitation learning policy performance by 38.75\% across 1,000 evaluations on two tasks, compared to models trained without lighting generation.
    Ablation studies are further conducted on individual modules of the proposed framework.
     
    \item \textbf{Downstream Applications. }
    We showcase generations on three additional environmental factors using geometric and material properties estimated by our framework.
\end{itemize}

\section{Related Work}
\subsection{Data Generation for Robotic Manipulation}
To address the robotic data bottleneck, recent efforts~\cite{yu2023scaling, mandi2022cacti, chen2023genaug, bharadhwaj2024roboagent} have focused on semantic augmentation of real-world images leveraging text-driven generative models~\cite{rombach2022high} to introduce texture variation, visual distractors, etc. For example, ROSIE~\cite{yu2023scaling} proposes changing object textures by first segmenting generation regions~\cite{minderer2022simple} and then performing text-guided image inpainting~\cite{wang2023imagen}. Rendering techniques~\cite{mildenhall2021nerf} have also been applied for viewpoint generation~\cite{chen2024rovi, zhang2024diffusion, zhou2023nerf}. Another line of work breaks long-horizon tasks into object-centric subtasks or manipulation skills and replays transformed demonstrations in simulation to generate new data from a limited number of examples~\cite{mandlekar2023mimicgen, jiang2024dexmimicgen, nasiriany2024robocasa}. 
In this paper, we focus on the underexplored problem of novel lighting generation on real-world human demonstrations for robotic manipulation.

\subsection{Inverse Rendering and Relighting}
Relighting a scene typically requires identifying and altering its properties to produce the intended lighting effect~\cite{azinovic2019inverse}.
Inverse rendering facilitates this by providing separate or joint estimations of geometry~\cite{eigen2015predicting}, material~\cite{li2018cgintrinsics}, and lighting~\cite{gardner2017learning} in a scene. Based on input requirements, inverse rendering can be categorized into single-view~\cite{li2022physically} and multi-view~\cite{yao2022neilf} methods. 
While methods such as DPI~\cite{lyu2023diffusion} and FIPT~\cite{wu2023factorized} produce high-fidelity relighting results, they rely on multi-view inputs for scene reconstruction and domain-specific datasets~\cite{park2020seeing}, making them incompatible with most existing robotic dataset camera setups.
 In this work, we present the first integration of single-view inverse rendering into robotic manipulation for explicit modeling of scene geometry and material properties.

\subsection{Latent Video Diffusion Models}
Latent diffusion models~\cite{rombach2022high} generate images by iteratively denoising Gaussian noise to approximate the target distribution. 
Latent video diffusion models~\cite{blattmann2023stable, blattmann2023align} extend this framework to the video domain typically by introducing temporal mixing layers on top of pre-trained image generation architectures. 
For example, Stable Video Diffusion~\cite{blattmann2023stable} extends Stable Diffusion~\cite{rombach2022high} by inserting temporal convolution and attention layers after each spatial layer and finetuning on curated video data. 
The extended temporal coherence combined with flexible conditioning (e.g., text, reference frames) enables applications such as camera-controlled video generation~\cite{he2024cameractrl}, multi-view synthesis~\cite{voleti2024sv3d}, and video prediction~\cite{wang2024driving}.
In this work, we adopt this structure for video-to-video translation, transferring lighting in real-world human demonstrations to new lighting conditions using physically-based relit frames as reference.

\begin{figure*}[t]
    \centering
    \includegraphics[width=0.9\linewidth]{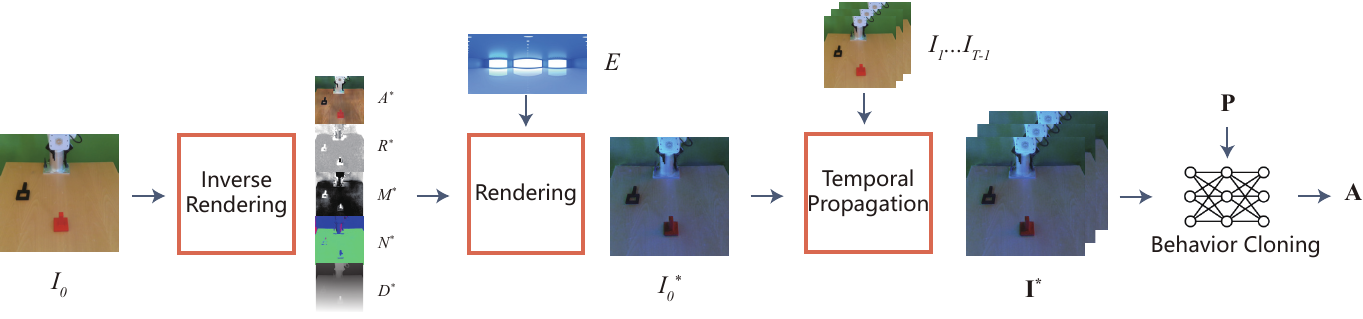}
    \caption{\small
    Given the first frame $I_0$ in a real-world human demonstration, the inverse rendering module estimates material ($A^*$, $R^*$, $M^*$) and geometric ($N^*$, $D^*$) properties. The rendering module uses these estimates and an environment map $E$ providing new lighting to produce a relit frame $I_0^*$. The new lighting is then propagated across the entire sequence $\mathbf{I}$, producing $\mathbf{I}^*$ and forming an generated episode ($\mathbf{I}^*$, $\mathbf{P}$, $\mathbf{A}$) for behavior cloning.}
    \label{fig:structure}
    \vspace{-1em}
\end{figure*}

\section{Methodology}
\subsection{Formulation and Overview}
Given one episode of real-world human demonstration ($\mathbf{I}$, $\mathbf{P}$, $\mathbf{A}$) recorded from a fixed viewpoint, where $\mathbf{I} \in \mathbb{R}^{H\times W \times 3 \times T}$ represents a sequence of $T$ RGB images, $\mathbf{P} \in \mathbb{R}^{p \times T}$ denotes proprioception states, and $\mathbf{A} \in \mathbb{R}^{a \times T}$ corresponds to actions.
Our goal is to transform $\mathbf{I}$ into $\mathbf{I}^*$ under new lighting, using physically based rendering (PBR) to ensure accurate light-material interactions.

\textbf{Overview.}
As shown in \textit{Fig.}~\ref{fig:structure} the proposed framework consists of three components:
(1) we begin by selecting the first frame $I_0$ from $\mathbf{I}$ and applying inverse rendering (\textit{Sec.}~\ref{sec:inverse_rendering}) to estimate the geometric and material properties of the scene depicted in $I_0$;
(2) the rendering module (\textit{Sec.}~\ref{sec:rendering}) then uses the estimated properties and an environment map $E$ to relight $I_0$, producing $I_0^*$; and
(3) the temporal propagation module (\textit{Sec.}~\ref{sec:temporal_propagation}) propagates the lighting from $I_0^*$ across the full image sequence $\mathbf{I}$, resulting in the final generated sequence $\mathbf{I}^*$. 
This forms an generated episode ($\mathbf{I}^*$, $\mathbf{P}$, $\mathbf{A}$), which is then used to train the imitation learning policy. 
The environment map $E$ has two use cases: (1) approximating the current lighting conditions to train a policy adapted to the current environment (\textit{Sec.}~\ref{sec:real_world_evaluation}), and (2) introducing diverse lighting to contribute to a lighting-invariant policy (\textit{Sec.}~\ref{sec:applications}).

\subsection{Single-Frame Inverse Rendering}
\label{sec:inverse_rendering}
Following~\cite{wang2025materialist}, we use the pre-trained network $\mathcal{P}$ to predict geometric and material properties of the scene depicted in $I_0$:
\begin{equation}
A_p, R_p, M_p, N_p, D_p = \mathcal{P}(I_0),
\label{eq:estimation}
\end{equation}
where $A_p$, $R_p$, and $M_p$ are predicted material properties (albedo, roughness, metallic), and $N_p$ and $D_p$ are predicted geometric properties (surface normal and depth). 
Examples of predicted properties can be found in \textit{Fig.}~\ref{fig:structure}. 
All predictions share the spatial resolution of $I_0$.
Final property estimates are derived by minimizing the following objective:
\begin{equation} 
A^*, R^*, M^*, N^*, D^* = \arg\min_{A_p, R_p, M_p, N_p, D_p} \mathcal{L}_p(I_0, I_p),
\end{equation}
where $I_p$ denotes the frame rendered using the predicted properties.
$\mathcal{L}_p = \mathcal{L}_{\textit{re}} + \delta \mathcal{L}_{\textit{cons}}$,
where $\mathcal{L}_{\textit{re}}$ denotes the reconstruction loss between the rendered frame $I_p$ and the original frame $I_0$, and $\mathcal{L}_{\textit{cons}}$ denotes the $L_1$ consistency loss the optimized and originally predicted properties, scaled by a factor $\delta$ set to 0.005 in most experiments.

\subsection{Single-Frame Relighting}
\label{sec:rendering}
\begin{figure}[h]
  \centering
  \begin{subfigure}[b]{0.22\linewidth}
    \centering
    \includegraphics[width=\textwidth]{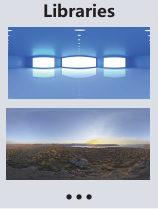}
    \vspace{-1.8em}
    \caption{\small}
    \label{fig:envmapA}
  \end{subfigure}
  \hspace{1em}
  \begin{subfigure}[b]{0.22\linewidth}
    \centering
    \includegraphics[width=\textwidth]{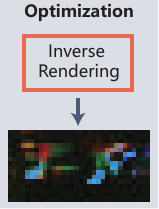} 
    \vspace{-1.8em}
    \caption{\small}
    \label{fig:envmapB}
  \end{subfigure}
  \hspace{1em}
  \begin{subfigure}[b]{0.22\linewidth}
    \centering
    \includegraphics[width=\textwidth]{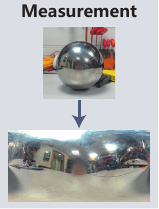} 
    \vspace{-1.8em}
    \caption{\small}
    \label{fig:envmapC}
  \end{subfigure}
  \vspace{-0.5em}
  \caption{\small (a) Environment maps from open-source HDRI libraries retrieved using keywords such as ``blue studio" and ``sunrise in the field". (b) Environment maps derived via optimization with inverse rendering, where resolution depends on the specific method. (c) Environment maps obtained by photographing a chrome (mirror) ball under multiple exposures and unwrapping the result.}
  \label{fig:envmap}
\end{figure}

To perform physically-based relighting, we render new lighting based on incoming radiance directions and modeled scene properties.
\paragraph{Target Light Sampling}
Given an environment map \( E \in \mathbb{R}^3 \) containing target lighting condition, the incoming radiance \( \lambda \) in direction \( \omega_i \) is sampled as:
\begin{equation}
\lambda(\omega_i) = E(\omega_i),
\label{eq:light_modelling}
\end{equation}
where \( E \) is obtained either from open-source HDRI liraries~\cite{PolyHavenHDRI} (\textit{Fig.}~\ref{fig:envmapA}), by 
optimization during inverse rendering~\cite{wang2025materialist} (\textit{Fig.}~\ref{fig:envmapB}) or by measuring the current lighting condition using graphics techniques~\cite{debevec2008rendering} (\textit{Fig.}~\ref{fig:envmapC}). 

\paragraph{Scene Modeling}The estimated properties from \textit{Sec.}~\ref{sec:inverse_rendering} are used to model the material appearance \( \mathcal{M} \) along direction \( \omega_i \) of the scene depicted in \( I_0 \), using the widely adopted Disney BRDF~\cite{burley2012physically}:
\begin{equation}
    \mathcal{M}(\omega_i)
 = (1 - M^*) f_{\textit{diffuse}}(A^*, R^*, N^*) + f_{\textit{specular}}(R^*, N^*), 
    \label{eq:material_modelling}
\end{equation}
where \( f_{\textit{diffuse}} \) and \( f_{\textit{specular}} \) represent the diffuse and specular reflection components of the Disney BRDF, respectively.

\paragraph{Scene Relighting}
Using the simplified rendering equation~\cite{kajiya1986rendering}, the generated frame \( I_0^* \) under new lighting from environment map $E$ is computed as:
\begin{equation}
    I_0^* = \int_{\Omega} \mathcal{M}(\omega_i) \lambda(\omega_i) (\omega_i \cdot N^*) \, \mathrm{d}\omega_i,
    \label{eq:rendering}
\end{equation}
where $\Omega$ is the hemisphere centered at surface normal $N^*$, containing all incoming directions $\omega_i$.
Note that once the property estimation for $I_0$ from \textit{Sec.}~\ref{sec:inverse_rendering} is completed, multiple generations can be applied using the same set of estimated properties.

\subsection{Temporal Propagation}
\label{sec:temporal_propagation}
\paragraph{Domain-Adapted Stable Video Diffusion}
To improve efficiency and maintain consistency in the generated sequences, we adopt Stable Video Diffusion (SVD)~\cite{blattmann2023stable} for its temporal modeling capabilities to propagate lighting from $I_0^*$ across the entire demonstration.
We formulate this as a video-to-video translation task: lighting from the original image sequence \( \mathbf{I} \) is transferred to generate the new sequence \( \mathbf{I}^* \), guided by the reference frame \( I_0^* \). This is defined as:
\begin{equation}
\mathbf{I}^* = \delta(I_0^*, \mathbf{I}_{\textit{input}}), \quad \mathbf{I}_{\textit{input}} = \text{Concatenate}(I_0^*, \mathbf{I}_{1:T-1}),
\label{eq:embedding}
\end{equation}
where \( \delta \) denotes the SVD model.

\begin{figure}[h]
  \centering
  \begin{subfigure}[b]{0.4\textwidth}
    \centering
    \includegraphics[width=\textwidth]{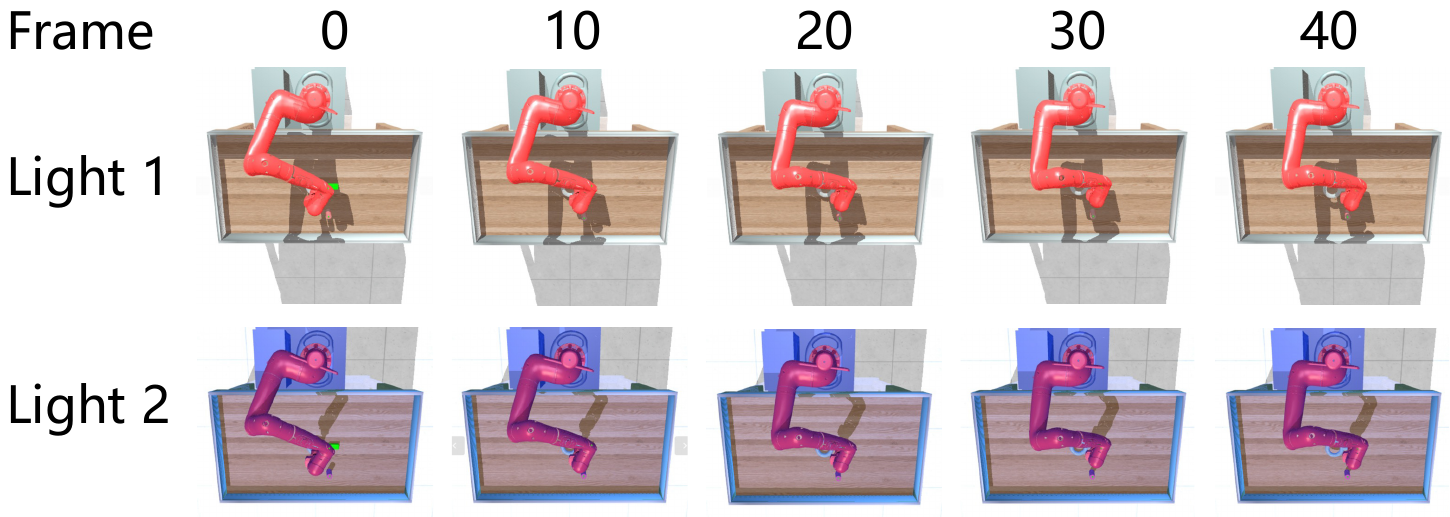}
    \vspace{-1.8 em}
    \caption{\small}
    \label{fig:synthetic_dataset}
  \end{subfigure}
  \begin{subfigure}[b]{0.4\textwidth}
    \centering
    \includegraphics[width=\textwidth]{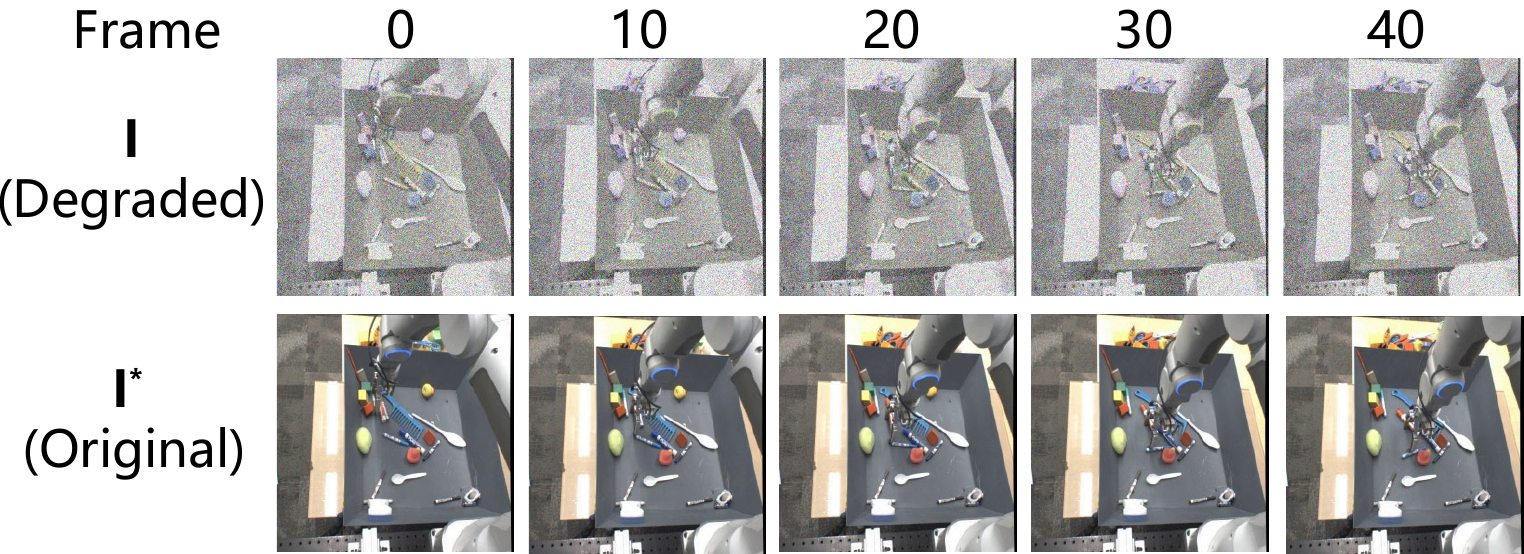} 
    \vspace{-1.8 em}
    \caption{\small}
    \label{fig:real_dataset}
  \end{subfigure}
  \vspace{-0.5em}
  \caption{\small(a) Examples of synthetic robot execution videos with lighting variations ($\mathcal{D}_1$). (b) Examples of real-world robot execution videos with visual degradation ($\mathcal{D}_2$). }
\end{figure}

\paragraph{Robotic Relighting Data Curation}
It’s important to note that the vanilla SVD underperforms in this task due to the absence of robotic elements like arms and grippers in its training data, resulting in a severely cartoonish appearance in generated robotic scenes.
Following SVD's original fine-tuning paradigm, we construct two datasets for different stages of fine-tuning:
\begin{itemize}
    \item Synthetic videos with lighting variation (\( \mathcal{D}_1 \), \textit{Fig.}~\ref{fig:synthetic_dataset}). We generate domain-specific relighting data for robotic manipulation using the Factor World~\cite{xie2024decomposing} benchmark, introducing lighting variation under identical task execution trajectories.  
    All 42 built-in scenes covering 19 tasks (e.g., door opening) are used, each captured with six camera views: one fixed top-down and five randomized within constrained azimuth, inclination, and radius in the range \( [-\pi/2, \pi/2] \) to ensure robot visibility.  
    Each scene includes 30 lighting conditions by sampling ambient and diffuse RGB values from \([25, 255]\).  
    Videos are rendered at \(512 \times 512\) resolution with 100 frames at 24 fps.  
    For fine-tuning, we randomly select pairs of videos showing the same task trajectories under different lighting, using one as the input \(\mathbf{I}\) and the other as the target sequence \(\mathbf{I}^*\).
    
    \item Real-world videos with visual degradation ($\mathcal{D}_2$, \textit{Fig.}~\ref{fig:real_dataset}). Since existing real-world robotic datasets lack perfectly paired original and relit videos under identical task execution trajectories, we manually create video pairs by applying visual degradation on original videos from RoboNet~\cite{dasari2019robonet} with 15 million video frames from 7 different robot platforms. 
    We apply random visual transformations by sampling brightness, contrast, and saturation scaling factors from $[0.2,1.9]$, and hue shifts from $[-0.5,0.5]$.
    The degraded robot execution video serves as the input $\mathbf{I}$, while the original video acts as the output $\mathbf{I}^*$. 
\end{itemize}

\paragraph{Fine-tuning} We fine-tune SVD in two stages. In Stage 1, we use synthetic videos from \( \mathcal{D}_1 \), which provide the main data for domain adaptation and reduce overfitting via controlled lighting variation. 
However, due to the non-physically-based lighting in Factor World, this stage lacks radiometric accuracy, motivating fine-tuning on real-world videos from $\mathcal{D}_2$ in Stage 2. 
Fine-tuning is performed on a pretrained 14-frame SVD model using a single H100 GPU for 8,000 steps for Stage 1 and 1,000 steps for Stage 2.
A limited number of steps are applied to Stage 2 due to the absence of ground-truth relit pairs.

\section{Experiments}
\subsection{Visual Lighting Quality Evaluation}
\begin{figure}[h]
    \centering    
    \includegraphics[width=0.9\linewidth]{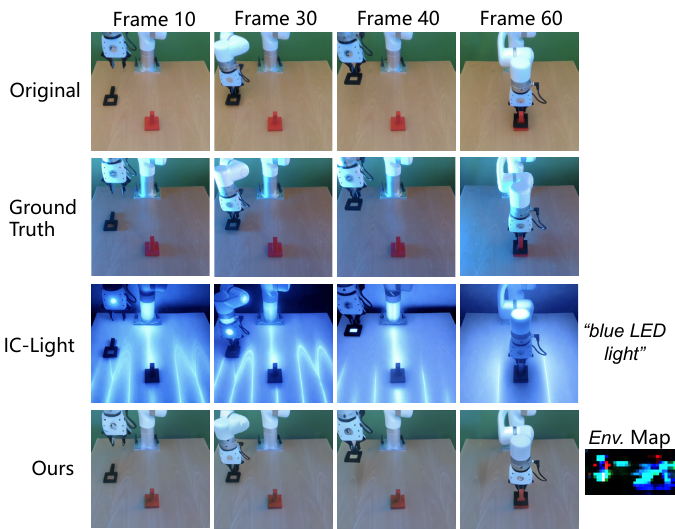}
    \begin{minipage}{\linewidth}
        \vspace{1 em}
        \centering
        \label{tab:texture_lighting_split}
        \resizebox{\linewidth}{!}{%
        \begin{tabular}{lcc|cc}
        \toprule
         Method & \textit{LPIPS} ($\downarrow$) & \textit{SSIM} ($\uparrow$) & \textit{Temporal LPIPS} ($\downarrow$) & \textit{Temporal SSIM} ($\uparrow$) \\
        \midrule 
        IC-Light & 0.5274 & 0.5359 & 0.121  & 0.898 \\ 
        Ours & \textbf{0.3269} & \textbf{0.7351} & \textbf{0.035}  & \textbf{0.978} \\
        \bottomrule
        \end{tabular}
        }
    \end{minipage}
    \caption{\small Qualitative (top) and quantitative (bottom) evaluation of lighting generation. Although exceeding the baseline, the lighting in our generated episodes appears more diffused compared to the ground truth, as indicated by weaker side shadows. This discrepancy mainly results from the rough environment map approximation ($32\times16$ resolution) without specialized graphics equipment.}
    \label{fig:qualitative_lighting}
    \vspace{-1em}
\end{figure}

We apply novel lighting generation to 10 randomly selected sequences (600 frames) from the real-world human demonstrations in \textit{Sec.}\ref{sec:real_world_evaluation} using our method and IC-Light~\cite{zhangscaling}, a widely recognized text-prompt-based relighting method.
Ground-truth episodes are recorded under a side-mounted blue LED panel light under the same task execution trajectories.
Two sets of metrics are adopted:
(1) \textit{LPIPS}~\cite{zhang2018unreasonable} and \textit{SSIM}~\cite{wang2004image} are computed between the ground-truth and relit image sequences to assess structural consistency and fidelity.
(2) \textit{Temporal LPIPS} and \textit{Temporal SSIM} are computed between consecutive frames in the relit sequences to assess temporal consistency.
Since IC-Light does not support video relighting, we apply it frame-by-frame using a fixed random seed and the prompt ``\textit{blue LED light}" for consistency.
We perform relighting using an environment map approximated from the ground truth episodes with~\cite{wang2025materialist}.
Qualitative and quantitative evaluation results can be found in  \textit{Fig.}~\ref{fig:qualitative_lighting}.
Our method performs better by preserving structural similarity with the original frames. 
The use of environment maps for target lighting provides precise control over relighting, making it more effective than textual prompts when approximating real-world data under varying lighting conditions.
The temporal propagation module further enhances visual consistency by reducing abrupt changes between frames. 
Beyond visual quality evaluation, IC-Light is also included in subsequent real-world comparisons in \textit{Sec.}~\ref{sec:real_world_evaluation}.

\begin{figure*}[t] %
  \centering
  \begin{subfigure}[b]{0.15\textwidth}
    \centering
    \includegraphics[width=\textwidth]{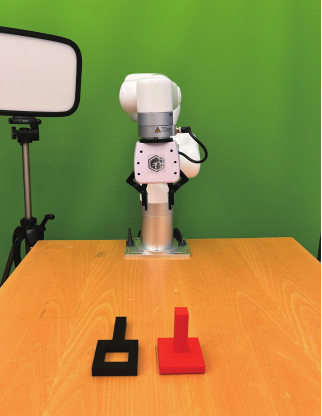}
    \vspace{-1.5em}
    \caption{\small}
    \label{fig:real_world_robot}
  \end{subfigure}
  \hspace{1em}
  \begin{subfigure}[b]{0.34\textwidth}
    \centering
    \includegraphics[width=\textwidth]{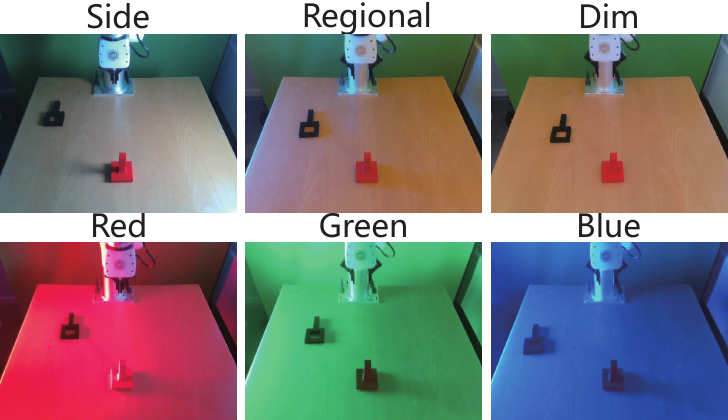} 
     \vspace{-1.5em}
    \caption{\small}
    \label{fig:real_world_lighting}
  \end{subfigure}
  \hspace{1em}
  \begin{subfigure}[b]{0.12\textwidth}
    \centering
    \includegraphics[width=\textwidth]{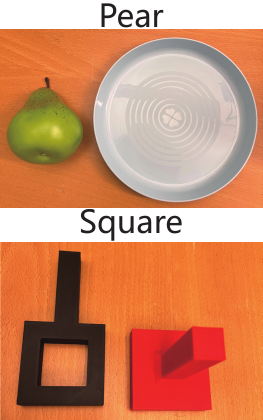} 
     \vspace{-1.5em}
    \caption{\small}
    \label{fig:real_world_task}
  \end{subfigure}
  \caption{\small(a) Experimental setup under \textit{Original} lighting. (b) The six lighting conditions used for policy evaluation under unseen lighting. (c) Objects used in the two manipulation tasks.}
  \vspace{-1em}
\end{figure*}

\begin{table*}[t]
    \centering
    \caption{\small Success rate ($\uparrow$) with different generation methods under unseen lighting. The best-performing policy is in bold, and the second-best is underscored.}
    \label{tab:generalization_gap}
    \resizebox{0.8\textwidth}{!}{ 
    \begin{tabular}{lcccc|cccc|cccc|cccc}
    \toprule
         \multirow{3}{*}{\textbf{Lighting}}& \multicolumn{4}{c}{\textit{Crop}} & \multicolumn{4}{c}{\textit{Jitter}} & \multicolumn{4}{c}{\textit{IC-Light}} & \multicolumn{4}{c}{\textit{Ours}} \\
        \cmidrule(lr){2-5} \cmidrule(lr){6-9} \cmidrule(lr){10-13} \cmidrule(lr){14-17} 
        &\multicolumn{2}{c}{\textit{Pear}} & \multicolumn{2}{c}{\textit{Square}}
        &\multicolumn{2}{c}{\textit{Pear}} & \multicolumn{2}{c}{\textit{Square}}
        &\multicolumn{2}{c}{\textit{Pear}} & \multicolumn{2}{c}{\textit{Square}}
        &\multicolumn{2}{c}{\textit{Pear}} & \multicolumn{2}{c}{\textit{Square}}\\
         \cmidrule(lr){2-3} \cmidrule(lr){4-5} \cmidrule(lr){6-7} \cmidrule(lr){8-9} \cmidrule(lr){10-11} \cmidrule(lr){12-13} \cmidrule(lr){14-15} \cmidrule(lr){16-17}
          & \textit{R} & \textit{PnP} & \textit{R} & \textit{PnP} &  
          \textit{R} & \textit{PnP} & \textit{R} & \textit{PnP} &
          \textit{R} & \textit{PnP} & \textit{R} & \textit{PnP} &
          \textit{R} & \textit{PnP} & \textit{R} & \textit{PnP}\\
        \midrule
        \textit{Original}& 1 & 0.60 & 1 & 0.85 &/ &/&/&/&/&/&/&/&/&/&/&/\\
        \midrule
        \textit{Side} & 0 & 0& 0 & 0& \textbf{0.3} & 0 & 0.15& 0&\underline{0.05}&0&\underline{0.2}&0&\underline{0.05} & 0 & \textbf{0.9} & \textbf{0.4}\\
        \textit{Regional} & 0.05 & 0 & 0&0 & \underline{0.45} & 0 & \underline{0.5} & 0 &0.10&0&0.15&0& \textbf{1.0} & \textbf{0.2} & \textbf{1.0} & \textbf{0.45}\\
        \textit{Dim} & \textbf{1.0} & \textbf{0.5} & \underline{0.95} & \underline{0.55} & 0.9 & 0.1 & 0.95 & 0.45 &0.20&0&0.25&0& \textbf{1.0} & \underline{0.4} & \textbf{1.0} & \textbf{0.8}\\
        \midrule
        \textit{Red} & 0 & 0 & 0 & 0& 0.3 & 0 & 0.1& 0&\underline{0.45}&0&\underline{0.3}&0& \textbf{0.6} & 0 & \textbf{0.9} & 0 \\
        \textit{Green} & 0.15 &0 & \underline{0.1} & 0 & 0& 0 & 0.05 & 0 &\underline{0.25}&0&\underline{0.1}&0& \textbf{1.0} & 0 & \textbf{1.0} & \textbf{0.1}\\
        \textit{Blue} & 0 & 0 & 0.05 &0 & 0 & 0 & \underline{0.45} & 0 &0.4&0&0.15&0& \textbf{0.85}& \textbf{0.05} & \textbf{0.95} & 0 \\
        \bottomrule
    \end{tabular}
    }
    \begin{tablenotes}
    \item 
    The vanilla BC-MLP with random cropping augmentation (\textit{Crop}) evaluated under \textit{Original} lighting is added as the reference of the task difficulty.
    \end{tablenotes}
    \vspace{-1em}
\end{table*}

\subsection{Real-world Evaluation}
\label{sec:real_world_evaluation}
This experiment examines whether imitation learning policies perform better under unseen lighting when trained on episodes that approximate such conditions.

\subsubsection{Implementation Details}
\paragraph{Lighting Setup}
White light (\textit{Original}, \textit{Fig.}~\ref{fig:real_world_robot}) used in most existing robotic manipulation datasets is adopted during expert human demonstration collection for policy training.
During evaluation, six lighting conditions are designed using a programmable RGB LED panel.
This includes two sets with different evaluation purposes:
(1) Daily-use conditions—\textit{Side} (strong shadows), \textit{Regional} (blue regional highlight), and \textit{Dim} (low-light);
(2) Artificial RGB lighting—\textit{Red}, \textit{Green}, and \textit{Blue}—with illuminance levels of 4,600 lux and RGB values (255, 0, 0), (0, 255, 0), and (0, 0, 255), respectively, designed to evaluate each policy under single-source lighting. These colors are chosen based on the principle of light transport linearity, which allows any colored illumination to be reconstructed as their linear combination.
Examples are shown in \textit{Fig.}~\ref{fig:real_world_lighting}.

\paragraph{Manipulation Tasks}
(1) \textit{Pear}, grasp a randomly placed plastic pear using a vacuum gripper and place it on a plate.
(2) \textit{Square}, grasp a randomly placed black square nut by its handle using a parallel jaw gripper and insert it into a red square peg.
To better assess performance under challenging lighting conditions, we report performance separately for the reaching (\textit{R}) subtask, which involves reaching the area of the pear or square nut, and the pick-and-place (\textit{PnP}) outcome, defined as successfully picking up the object and placing it in the corresponding plate or peg.

\paragraph{Baselines}
We perform behavior cloning with ResNet18~\cite{he2016deep} and a Multi-Layer Perceptron (BC-MLP) to map features to actions. 
This architecture is chosen for its widespread use~\cite{dasari2023unbiased, mandlekar2021matters} and simplicity, providing a clean testbed to evaluate the impact of generated data. 
Training is conducted on a 7-DoF UFactory XArm7 robot observed by a single externally mounted RGB camera.
We compare four baselines: 
(a) \textit{Crop}, vanilla BC-MLP with random cropping, trained on 200 human demonstration episodes under \textit{Original} lighting; 
(b) \textit{Jitter}, random cropping plus color jitter, trained on the same 200 episodes; 
(c) \textit{IC-Light}, trained on the original 200 episodes plus generated episodes approximating unseen lighting conditions with IC-Light; 
and (d) \textit{Ours}, trained on the original 200 episodes plus generated episodes from our method. 
For \textit{IC-Light} and \textit{Ours}, 10 demonstrations are randomly selected and relit with six environment maps matching the six evaluation conditions (derived through inverse rendering), yielding 260 training episodes per task. 
Color jitter is included as a baseline due to its ability to mitigate color and illumination variation.

\paragraph{Evaluation Protocol}
For each combination of task, lighting condition, and method, we conduct 20 real-world evaluations and report their success rates.
\textit{Pear} (\textit{Fig.}~\ref{fig:firstA}) and \textit{Square} (\textit{Fig.}~\ref{fig:real_world_robot}) differ slightly in background, and the wrist camera is included only for automatic evaluation, not for training purposes.
All baselines within a task are evaluated in the same environment.

\subsubsection{Results}
\textit{Tab.}~\ref{tab:generalization_gap} presents the success rates of BC-MLP under six unseen lighting conditions across 1,000 real-world evaluations. 
Averaged over both subtasks across the two tasks, our method outperforms \textit{Crop} by 38.75\%, \textit{Jitter} by 33.13\% and \textit{IC-Light} by 41.88\%.
In the daily-use lighting set, \textit{Crop} shows a major drop in performance, except under \textit{Dim}. 
While \textit{Jitter} performs reasonably on the reaching subtask, it consistently fails in the final pick-and-place outcome.  
\textit{IC-Light} exhibits reduced performance on daily-use lighting but better performance on RGB lighting, possibly because it generates episodes that align more closely with RGB lighting.
\textit{Ours} improves performance across both subtasks under these conditions.
In the artificial RGB lighting set, both \textit{Crop} and \textit{Jitter} fail in most cases, even on the reaching subtask. Interestingly, \textit{Jitter} displays strong preferences for different RGB lighting in the two tasks, likely due to its use of global color shifts. 
Our method maintains strong reaching performance but struggles with pick-and-place in these more extreme lighting conditions.
Common failure modes for \textit{Crop}, \textit{Jitter} and \textit{IC-Light} involve moving directly to the plate or peg without interacting with the pear or nut. 
Failures in \textit{Ours} often involve reaching the pear or nut with the gripper open and hovering just above the object, without executing a pick action.
This may relate to the altered object appearance under high-intensity lighting (4,600 lux), where high reflectivity could disrupt perception. See the limitations section for further details and the supplementary video for examples of common failure modes across different methods.

\subsection{Ablation Study}
We provide ablation study on individual modules of the proposed framework:
Inverse Rendering (\textit{Sec.}~\ref{Sec:ablation_material}), Temporal Propagation (\textit{Sec.}~\ref{sec:ablation_SVD}, \textit{Sec.}~\ref{sec:ablation_efficiency}) and Behavior Cloning (\textit{Sec.}~\ref{sec:ablation_num_episodes}).

\subsubsection{Material Property Ablation}
\label{Sec:ablation_material}
As shown in \textit{Tab.}~\ref{tab:material_ablation}, we ablate material properties by individually masking estimated albedo, roughness, and metallic to 0.5 (originally in [0, 1]) and evaluate the trained policy's success rate (\textit{R}/\textit{PnP}) over 10 trials on task \textit{Square} under \textit{Dim}. 
Policy trained on the original estimates is included for comparison. 
Results suggest albedo and roughness are critical for policy performance, while metallic has minimal effect, likely due to the non-metallic object. 
Masking any property leads to reduced performance.
\begin{table}[h]
    \vspace{-1em}
     \centering
     \caption{\small Material property ablation.}
    \resizebox{0.7\linewidth}{!}{ 
    \begin{tabular}{l|ccc}
    \toprule
         \textit{Original}& \textit{Albedo} & \textit{Roughness} & \textit{Metallic} \\
        \midrule
          1.0/0.8 & 0.8/0 & 1.0/0 & 1.0/0.4 \\
        \bottomrule
    \end{tabular}
    }
    \vspace{-1em}
    \label{tab:material_ablation}
\end{table}

\subsubsection{SVD Fine-tuning Data Ablation}
\label{sec:ablation_SVD}
\begin{figure}[h]
    \vspace{-1em}
    \centering
    \includegraphics[width=0.5\linewidth]{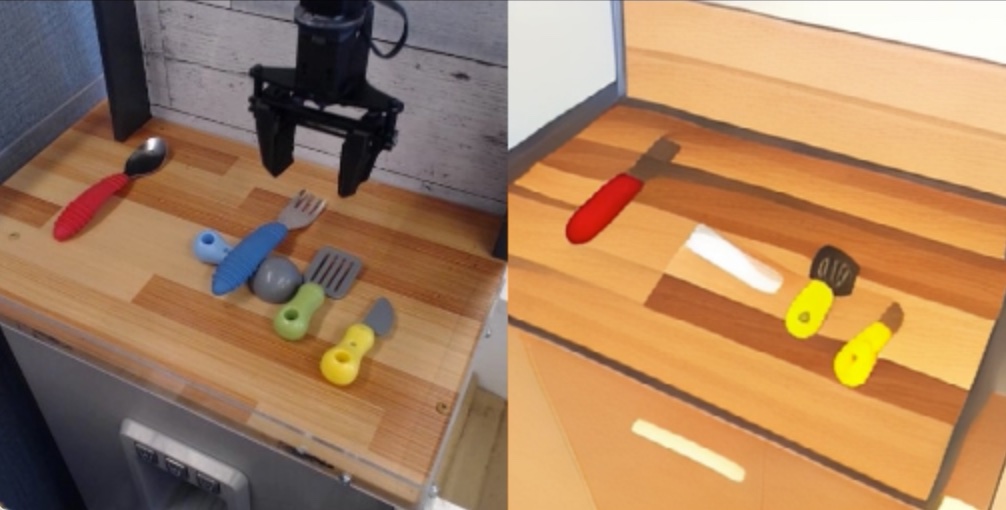}
    \caption{SVD fine-tuning data ablation.}
    \label{fig:ablation_study}
    \vspace{-1em}
\end{figure}

The fine-tuning data for SVD consist of two sources: synthetic robot execution videos, which provide varied lighting across tasks (\( \mathcal{D}_1 \)), and real-world execution videos, which include visual degradation to improve radiometric accuracy (\( \mathcal{D}_2 \)). 
In \textit{Fig.}~\ref{fig:ablation_study}, we show an example frame generated by SVD when fine-tuned only on \( \mathcal{D}_1 \). 
The result exhibits low radiometric accuracy and a noticeably cartoonish appearance.

\subsubsection{Temporal Propagation Ablation}
\label{sec:ablation_efficiency}
\textit{Tab.}~\ref{tab:efficiency} shows the comparison of relighting time: IC-Light, frame-wise inverse rendering (\textit{Ours (w/o SVD)}), and the proposed first frame inverse rendering + SVD (\textit{Ours (w/ SVD)}). With SVD, after a one-time 10-minute inverse rendering, relighting each episode takes only 20 seconds. The efficiency gain becomes greater when relighting an episode with multiple lighting, e.g. 6 relightings take just 12 minutes: $10+ (20/60) \times 6$.
\begin{table}[h]
    \centering
    \caption{\small Time efficiency comparison.}
    \resizebox{\linewidth}{!}{
    \begin{tabular}{l|c|c|c}
    \toprule
    Methods & \textit{1 Frame} & \textit{1 Episode (61 Frames)}  & \textit{6 Episodes} \\
    \midrule
    \textit{IC-Light} & 10 s & 10 min & 1 hour  \\
    \textit{Ours (w/o SVD)} & 10 min  & 10 hours & 60 hours \\
    \textit{Ours (w/ SVD)} & 10 min  & 10 min + 20 s & \textbf{12 min}    \\
    \bottomrule
    \end{tabular}
    }
    \label{tab:efficiency}
    \vspace{-1em}
\end{table}

\subsubsection{Impact of the Number of Generated Episodes for Policy Training}
\label{sec:ablation_num_episodes}
We select 10, 25, 75, and 100 episodes from 200 demonstrations and generate each with six lighting conditions.
Evaluation is performed in single-source \textit{Red} light to isolate generation impact.
Reaching success rates on \textit{Square} task, averaged over 10 trials.
As shown in \textit{Tab.}~\ref{tab:num_episodes}, performance is relatively unaffected by the number of generated episodes, with no signs of overfitting.
To save computation, we use the minimal setting of 10 generated episodes for policy training.
\begin{table}[h]
     \centering
     \vspace{-1em}
     \caption{\small Impact of the number of generated episodes.}
    \resizebox{0.6\linewidth}{!}{ 
    \begin{tabular}{lcccc}
    \toprule
         \textit{Num. Episodes}& \textit{10} & \textit{25} & \textit{75} & \textit{100} \\
        \midrule
         \textit{Success Rate (R)} &0.9 & 0.8 & 0.8 & 1.0 \\
        \bottomrule
    \end{tabular}
    }
    \label{tab:num_episodes}
\end{table}

\section{Applications}
\label{sec:applications}
\begin{figure}[h]
    \vspace{-1em}
    \centering
    \includegraphics[width=0.7\linewidth]{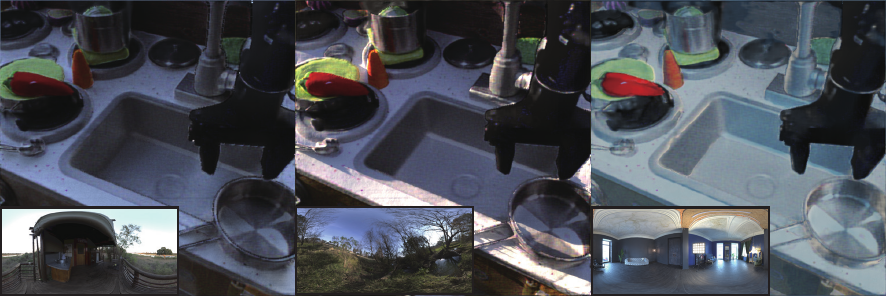}
    \caption{\small Examples of lighting generation with different environment maps on existing open-source dataset.}
    \label{fig:lighting}
    \vspace{-1em}
\end{figure}

Through scene decomposition via inverse rendering, we extend our framework to generate three additional environment factors (background, object texture, and distractors) beyond lighting (\textit{Fig.}~\ref{fig:lighting}) on existing datasets, as demonstrated on BridgeData v2~\cite{walke2023bridgedata}.

\subsection{Background Generation}
\label{sec:background_generation}
\begin{figure}[h]
    \centering
    \includegraphics[width=\linewidth]{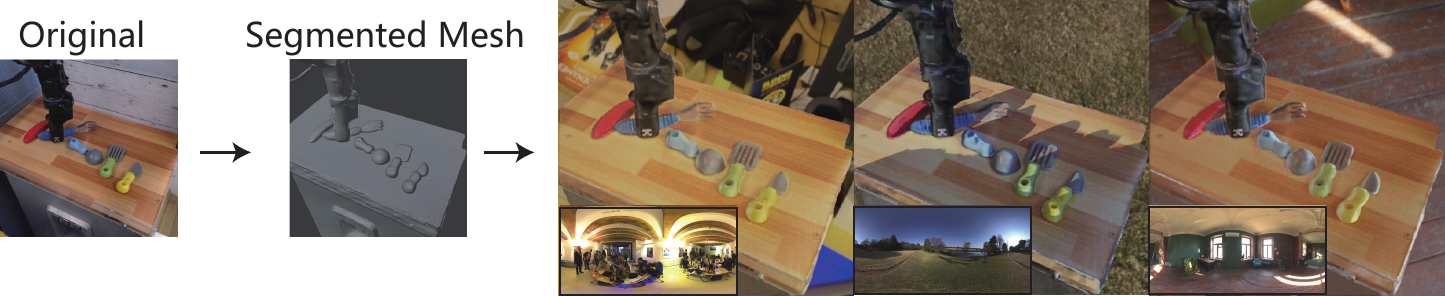}
    \caption{\small Examples of background generation with different environment maps from~\cite{PolyHavenHDRI}.}
    \label{fig:background}
    \vspace{-1em}
\end{figure}

Background appearance is tied to illumination; therefore, we render segmented scene geometry with different environment maps to create new backgrounds with corresponding lighting.
Scene geometry is reconstructed by triangulating the depth map \( D^* \) from \textit{Sec.}~\ref{sec:inverse_rendering} into a mesh \( \Lambda \):
\begin{equation}
\Lambda = \text{Triangulate}\left(K^{-1} [x\;\; y\;\; 1]^\top D^*(x, y)\right),
\label{eq:depth_to_mesh}
\end{equation}
where $K$ is a fixed intrinsic matrix of a pinhole camera, and $D^*(x, y)$ is the depth at pixel $(x,y)$.
We then project 2D segmentation masks~\cite{kirillov2023segment} onto the mesh $\Lambda$ from the default viewpoint to isolate and segment the robot arm and task area. 
The segmented mesh is then rendered with various environment maps to produce new backgrounds and lighting.
Examples are shown in \textit{Fig.}\ref{fig:background}.

\subsection{Object Texture Generation}
\label{sec:object_texture}
\begin{figure}[h]
    \centering
    \includegraphics[width=0.7\linewidth]{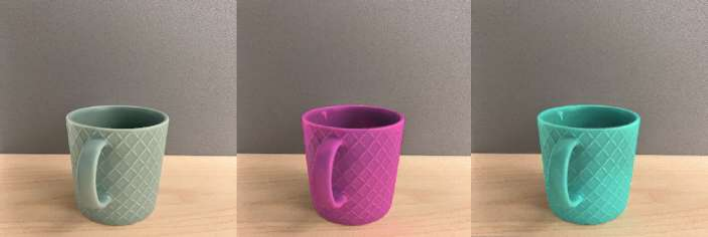}
    \caption{\small Examples of texture generation via albedo adjustment.}
    \label{fig:texture}
    \vspace{-1em}
\end{figure}

Robotic tasks like grasping and pushing rely on consistent object geometry and frictional properties ~\cite{mahler2016dex}.
We address this by applying object texture generation with preserved visual roughness and geometry.
Following the mesh segmentation in \textit{Sec.}~\ref{sec:background_generation}, we segment the object mesh and render with altered material properties. 
Specifically, albedo \( A^* \) sets the base color, roughness \( R^* \) controls surface scattering, and metallic \( M^* \) defines the metallic effect.
By adjusting only the albedo \( A^* \) while keeping roughness \( R^* \) and metallic \( M^* \) fixed, and reapplying \textit{Eq.}~\ref{eq:material_modelling}, we achieve object texture generation with consistent visual roughness and geometry.
Examples are shown in \textit{Fig.}~\ref{fig:texture}.

\subsection{Distractor Placement in Cluttered Environments}
\label{sec:distractor}
\begin{figure}[h]
    \centering
    \includegraphics[width=\linewidth]{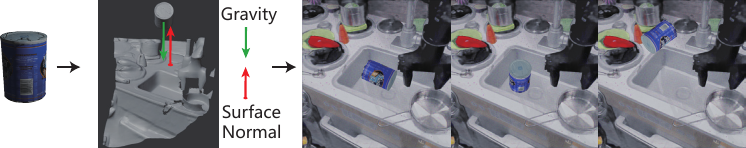}
    \caption{\small Examples of gravity-based distractor placement, with distractor mesh sourced from~\cite{calli2015benchmarking}.}
    \label{fig:distractor}
    \vspace{-1em}
\end{figure}

Through the meshes of the scene and distractors, we conduct physically plausible distractor placement in Blender~\cite{blender}.
As shown in \textit{Fig.}~\ref{fig:distractor}, we first align the surface normal with Blender's default gravity axis (Z-axis). This requires the user to manually select a mesh face whose surface normal is parallel to the gravity direction using the Blender API.
A random position within the bounding box of $\Lambda$ (\textit{Eq.}~\ref{eq:depth_to_mesh}) is sampled, and the distractor mesh (assigned a pseudo-mass of 1kg at its geometric center) is dropped to simulate its motion under gravity. 
The final placement of the distractor is determined by its stable resting state following the gravity-based simulation. 
The full simulation process is available in the supplementary video.

\section{Limitations and Conclusion}
\subsection{Limitations and Future Work}
\begin{figure}[h]
    \centering
    \includegraphics[width=\linewidth]{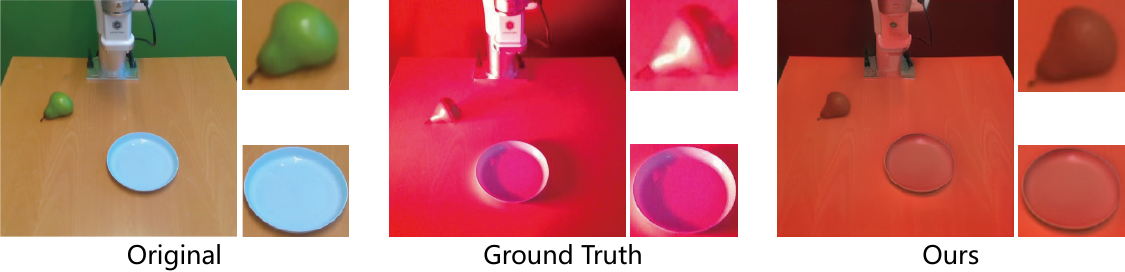}
    \caption{Failure cases observed under high-intensity lighting.}
    \label{fig:limitation}
    \vspace{-1em}
\end{figure}

Although we provide physically-based lighting generation by simulating light–material interactions, the absence of an environment map that perfectly reconstructs the observed lighting reduces performance. 
An example is shown in \textit{Fig.}~\ref{fig:limitation}, where under high-intensity red LED lighting (4600 lux), the plastic pear and plate exhibit strong reflections not captured by our generation. 
The discrepancy arises from the side-mounted LED producing strong directional lighting, while the approximated environment map ($32 \times 16$ resolution) fails to capture this and instead provides mostly diffuse illumination with a shifted color tone. 
Capturing current lighting with graphical equipment~\cite{debevec2008rendering} to generate accurate environment maps could mitigate this issue. Apart from above, our paper focuses on improving policy performance under unseen lighting by generating data that approximates such conditions, rather than developing lighting-invariant policies that function across all diverse lighting. 
Future work will aim to systematically study how different lighting conditions affect policy performance, combined with lighting generation techniques, toward developing strategies that enable lighting-invariant policies.

\subsection{Conclusion}
In this paper, we propose the first framework that leverages physically-based inverse rendering for novel lighting generation on existing real-world human demonstrations.
By decomposing robotic scenes into geometric and material properties, we perform lighting generation on a single frame and propagate it across the entire demonstration using finetuned Stable Video Diffusion.
We validate our framework through qualitative and quantitative visual quality evaluations, 1,000 real-world trials under six varied lighting conditions using a 7-DoF robot, and ablation studies on individual modules of the proposed framework.
We further showcase generations on three additional visual variations enabled by our framework. 
The code for our framework will be publicly released to support future research in physically-based data generation for robotic manipulation.

\bibliographystyle{IEEEtran}
\balance
\bibliography{citations}

\end{document}